# A MULTI-TASK DEEP LEARNING FRAMEWORK FOR BUILDING FOOTPRINT SEGMENTATION

*Burak Ekim* [1,3] *and Elif Sertel*[*] [1,2]

[1] Center for Satellite Communications and Remote Sensing,
Istanbul Technical University, Istanbul, 34469, Turkey
[2] Geomatics Engineering Department, Istanbul Technical University, Istanbul, 34469, Turkey,
[3] Institute of Informatics, Satellite Communication and Remote Sensing Program,
Istanbul Technical University, Istanbul, 34469, Turkey

## ABSTRACT

The task of building footprint segmentation has been well-studied in the context of remote sensing (RS) as it provides valuable information in many aspects, however, difficulties brought by the nature of RS images such as variations in the spatial arrangements and in-consistent constructional patterns require studying further, since it often causes poorly classified segmentation maps. We address this need by designing a joint optimization scheme for the task of building footprint delineation and introducing two auxiliary tasks; image reconstruction and building footprint boundary segmentation with the intent to reveal the common underlying structure to advance the classification accuracy of a single task model under the favor of auxiliary tasks. In particular, we propose a deep multi-task learning (MTL) based unified fully convolutional framework which operates in an end-to-end manner by making use of joint loss function with learnable loss weights considering the homoscedastic uncertainty of each task loss. Experimental results conducted on the SpaceNet6 dataset demonstrate the potential of the proposed MTL framework as it improves the classification accuracy greatly compared to single-task and lesser compounded tasks.

***Index Terms*—** Deep learning, Fully-convolutional networks, Multi-task learning, Remote sensing, Semantic segmentation,

## 1. INTRODUCTION

Building footprint segmentation, one of the most disseminated tasks of remote sensing (RS) image segmentation, aims at delineating building footprints from an overhead perspective which is of great value in various applications such as urban sprawl monitoring, city and regional planning, smart city applications, and disaster management. In recent years, RS based applications draw advantage from the advent of deep learning-powered techniques by exploiting the ever-growing data volumes in virtue of developing the earth observation industry, and more accessible high-resolution RS images [1]. Yet, the task of semantic segmentation, especially building footprint segmentation, hosts various challenges owing to the characteristic of RS images, such as site-specific cultural and social-economical dynamics of the world; material of construction (e.g. tile and brick), and roof-top structure of the building, and remains to be further investigated. Further, the delineation of building boundaries and preserving the form of footprints such as edge and boundary characteristics also obstruct the way of creating less problematic segmentation maps [2]. MTL aims to optimize many tasks simultaneously in a unified framework to reveal the common underlying structure, with an aspiration to improve the single-task performance [3]. From the machine learning point of view, MTL can be considered as a form of inductive transfer that helps to improve the model by introducing an inductive loss, thus acts as a regularizer and enables solutions that often generalize better [4]. Nevertheless, the superiority of MTL over single-task learning is not steady in that it requires careful selection of tasks and weighting strategies by making it a continuing research direction [5]. Weighting strategies refer to obtaining the individual tasks that contribute to the overall objective function and contain diversified methods; for example, the dynamic weight average (DWA) [6] makes use of the rate of loss change for each task and converts this quantity into a weight for each task. Similarly, GradNorm [7] utilizes dynamically tuning gradient magnitudes to define the loss weights. In [8], the authors derive the loss weights depending on the homoscedastic uncertainty of each task, which also adopted in this paper. There are several MTL segmentation research that exists in the literature. In [9], the authors aimed to harness the image reconstruction task to improve segmentation accuracy. Similarly, in [2], additional boundary information is deployed with the aim of preserving the boundary information. In, [10] authors deployed MTL logic to cope with stereo DSM filtering and roof type classification, introducing surface normal objective along with conditional adversarial objective to the DSM filtering which was considered a reconstruction task. In [11], human settlement extent regression and local climate zone classification tasks found to be correlated and provide hints for each other.

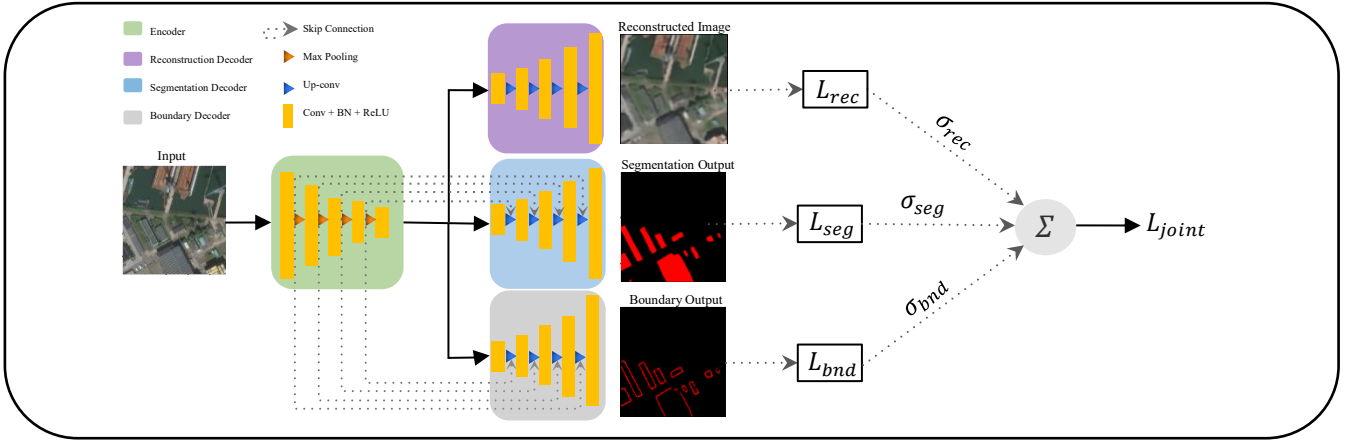

**Fig. 1:** Illustration of the proposed framework.

In this paper, we propose an MTL-guided framework to simultaneously address building footprint segmentation as the main task; building boundary extraction, and image reconstruction as auxiliary tasks for very high-resolution RS images. Further, we provide an implementation of homoscedastic uncertainty-powered logic to learn the auxiliary losses contributions to the overall objective function. We show that harnessing from auxiliary tasks yields useful information for semantic segmentation which manifests itself with the aid of the proposed unified framework which outperforms the single-task model.

## 2. METHODOLOGY

U-Net [12], a fully convolutional network-based architecture, aims at making a pixel-wise, dense, prediction of an input image with the use of successive encoder and decoder blocks. To extend the single-task to a multi-task scheme, multiple decoder branches need to be constituted where each decoder outputs distinct outputs of tasks, as shown in Figure 1.

### 2.1. MULTI-TASK DESIGN

Multi-task learning is the process of optimizing a model with respect to more than one objective. The traditional and straightforward way is to simply add the individual losses together, favorably in the form of weighted sum, which would give relativistic importance to each loss function according to usefulness [8]. However, this approach becomes tedious as the number of tasks increases, since it requires an extensive number of experiments which consume resources time, and computational power. To this end, we apply a logic that allows learning the contribution of each loss function.

### 2.2. Learning with Homoscedastic Uncertainty

As the number of tasks involved in the MTL process increases, tuning the individual loss weights by hand becomes a prohibitive and not sustainable approach as stated by the previous works, and also by the empirical results obtained in this paper, as illustrated in Figure 2. To this end, motivated by [8], we make use of homoscedastic uncertainty of each task to learn the contributions of the individual losses to the overall objective function, instead of manually tuning the weights. This way, the joint loss function is defined as,

$$L_{joint} = \sum_{t=1}^{T} \omega_t \cdot L_t \qquad (1)$$

Where $\omega_t$ indicates the weight (or contribution) of the loss function to overall objective function, in which $t = 1,2,\cdots,T$. In order to prevent the individual objective function result in trivial solutions for $\omega_t = 0$ the constrained optimization problem is introduced by adopting regularization term $r_t$ to each task, with regard to limit the optimization space. After the implicit definition of each objective function, as: $L_{seg}, L_{bnd}, L_{rec}$, joint loss function appears to be,

$$L_{joint} = \omega_{seg}L_{seg} + r_{seg} + \omega_{bnd}L_{bnd} + r_{bnd} \\ + \omega_{rec}L_{rec} + r_{rec} \qquad (2)$$

Optimizing $L_{joint}(W, \sigma_{seg}, \sigma_{bnd}, \sigma_{rec})$ with respect to the learnable parameters, (model parameters along with individual loss weights) by utilizing task-specific weighting based on homoscedastic task uncertainty leads to the overall joint loss to minimize,

$$= \frac{1}{\sigma_{seg}^2}L_{seg}(W) + \log \sigma_{seg} + \frac{1}{\sigma_{bnd}^2}L_{bnd}(W) + \\ \log \sigma_{bnd}^2 + \frac{1}{2\sigma_{rec}^2}L_{rec}(W) + \log \sigma_{rec}^2 \qquad (3)$$

For the sake of numerical stability, exponential mapping of $\sigma_t$ is employed and initial values are selected as 0, despite the fact that this technique is robust to the selection of initial values, as stated in [8].

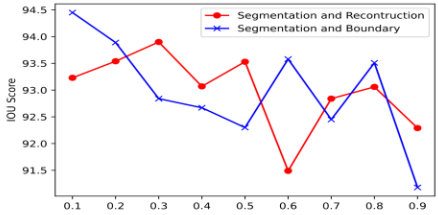

**Fig.2.** Tuning the loss weights by hand. Weight values indicate the contribution of the particular auxiliary task to the overall objective function.

## 3. EXPERIMENTS

### 3.1. Dataset

We use the open-source SpaceNet 6: Multi-Sensor All-Weather Mapping [13] dataset. This is a multi-sensor dataset with two modalities; optical and SAR. In this paper, we consider optical modality, particularly, pan-sharpened RGB image product. Data is acquired from the WorldView 2 satellite with a 0.5-meter spatial resolution and covers the area of Rotterdam, the Netherlands. The dataset focuses on building footprint segmentation and is annotated with over 48,000 unique building footprint labels, making it suitable for the targeted task in this paper.

### 3.2. Implementation Details

The proposed framework takes input image, ground truth map, and boundary extracted ground truth map as an input and outputs; (1) segmentation map, (2) boundary segmentation map, and (3) reconstructed input image. The objectives aimed to optimize in this paper are, cross-entropy for semantic segmentation and boundary extraction tasks and mean absolute error (MAE or L1 loss) for the reconstruction task. We use the canny edge detection algorithm with a radius of 3, together with dilation to extract the boundaries from whole ground truth masks. During experiments, the input images and corresponding masks are randomly cropped by 512 x 512 patches as network inputs. The dataset is partitioned by a 7:2:1 ratio for train, validation, and test set, respectively. We apply data augmentation to increase the dataset size artificially by deploying the following transformations: horizontal-vertical flip, additive Gaussian noise, perspective, random brightness, gamma, contrast, blur. Our encoder-decoder type network architecture is based on U-Net [12], with five encoder and following decoder blocks with the filter sizes of (256, 128, 64, 32, 16), as depicted in Figure 1. All decoders are the same except for the reconstruction task in which skip connections are neglected with an intent to force the model not to peek at the latent space information, consequently creating the reconstructing image in a semi-aided way. ResNet101 [14] is used as a backbone for the feature encoder. For all experiments, we use a learning rate of $2.5 \times 10^{-3}$ and train using SGD with the momentum of 0.9 and weight decay of $10^4$. We conduct the experiments using PyTorch. As hardware, we use GeForce RTX 2080 Ti.

### 3.3 Experimental Results

Motivated by the empirical results shown in Figure 2, we can conclude that the weight of each task has a substantial impact on model performance. The need for learning the loss weights further increases with the increasing number of tasks included in the joint loss function. This scenario is the main motivation behind employing homoscedastic uncertainty powered logic in order to learn the contribution of each task's loss to the overall loss function in a simultaneous way.

**Table 1**. Quantitative results on test set. S: Semantic Segmentation, R: Image Reconstruction, B: Boundary Extraction, P: Post Processing.

| Experiments | IOU Score | F1 Score |
| --- | --- | --- |
| S | 92.03 | 95.82 |
| S + R | 93.15 | 96.12 |
| S + B | 94.05 | 96.47 |
| S + B + R | 96.97 | 98.45 |
| (S + B + R) with P | 97.45 | 98.86 |

Table 1 presents the comparison of single-task learning and multi-task learning in terms of F1 score and IOU score. From this table, it is obvious that taking auxiliary tasks, either boundary extraction or reconstruction task into consideration along with the main task increases the model performance. Such that, image reconstruction coupling with semantic segmentation advances the performance by 1.12%, in a similar way, making use of boundary information advances the performance by 2.02%. Besides, as a post-processing technique, fusing the segmentation and boundary tasks and applying opening operation afterward advance the performance further, as shown in the last row of Table 1. Thus, in the light of experimental results conducted, one can conclude that complementary information can be learned from auxiliary tasks. Particularly, reconstructing the input image helps in improving the segmentation performance as it aims at regenerating the image that will be used for creating a segmentation map. Moreover, learning to extract the boundaries of the segmentation map is also beneficial as the model learns the boundaries of buildings, thus guides the model to learn the boundary information as shown in Figure3.

## 4. CONCLUSION

In this paper, we proposed a three-output MTL logic for building footprint segmentation approach based on the intuition that semantic segmentation could benefit from the boundary extraction and image reconstruction auxiliary tasks in a joint optimization scheme. Since tuning the individual losses contribution to the overall objective function by hand is time-consuming and requires numerous experiments to be conducted (Figure 2), we deployed a learnable loss weight logic. Furthermore, we applied postprocessing by fusing segmentation and boundary outputs which led to an increase in accuracy. Results show that the proposed framework improves the segmentation performance by 5 % compared to a single output building footprint segmentation task.

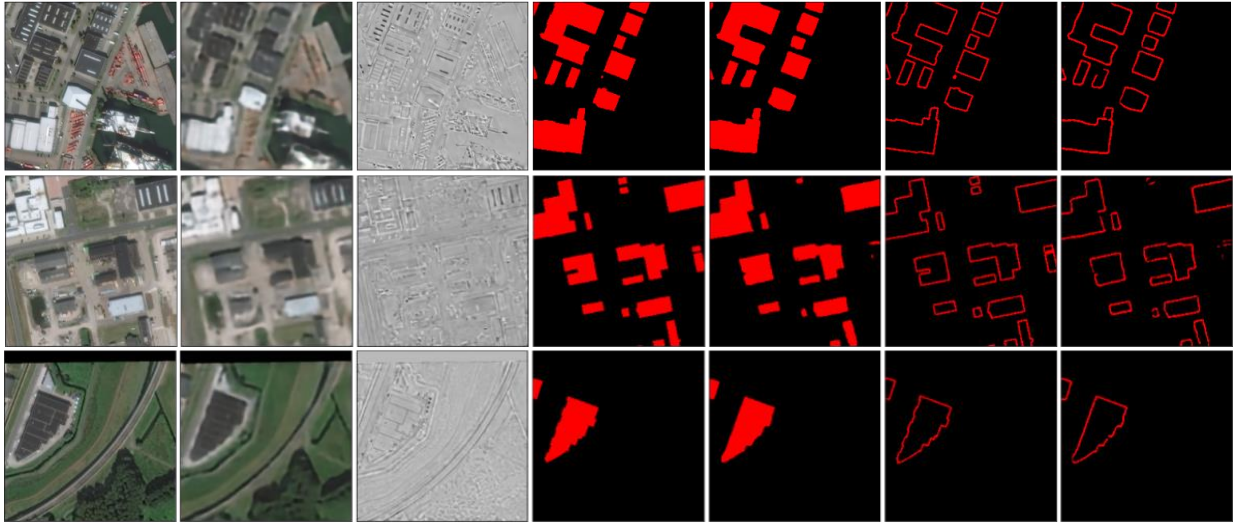

**Fig. 3**. Qualitative results on S+B+R model. From left to right: original image, reconstructed image, difference map, ground truth for segmentation, output segmentation map, ground truth for boundary extraction, output boundary map.